\documentclass[conference]{IEEEtran}
\usepackage{cite}
\usepackage{amsmath,amssymb,amsfonts}
\usepackage{graphicx}
\usepackage{textcomp}
\usepackage{xcolor,soul}
\usepackage{booktabs}
\usepackage{dsfont}

\begin{document}

\title{Are Spiking Neural Networks Harder to Steal? The Role of Stochastic Encoding in Model Extraction}

\author{\IEEEauthorblockN{Ayana Moshruba}
\IEEEauthorblockA{\textit{Department of Electrical and Computer Engineering} \\
\textit{George Mason University}\\
Fairfax, Virginia, USA \\
amoshrub@gmu.edu}
\and
\IEEEauthorblockN{Maryam Parsa}
\IEEEauthorblockA{\textit{Department of Electrical and Computer Engineering} \\
\textit{George Mason University}\\
Fairfax, Virginia, USA \\
mparsa@gmu.edu}
}

\maketitle

\begin{abstract}
Machine learning models deployed on edge devices are vulnerable to model stealing attacks, in which an adversary replicates a model using only its predictions. Existing defenses add mechanisms that are costly on resource-constrained hardware. Spiking neural networks (SNNs) offer energy-efficient inference at the edge and have been shown to leak less information about their training data than artificial neural networks (ANNs). This work investigates whether SNNs also resist model stealing and identifies the source of any such resistance. We evaluate four model stealing attacks against SNN and ANN models on three image datasets and introduce a metric that accounts for the stochastic outputs of SNNs. SNNs exhibit greater resistance than ANNs: on average, surrogate models extracted from SNNs agree with their targets 38\% less often outside the training distribution, and targeted adversarial examples crafted on these surrogates transfer 54\% less often. This resistance, however, arises mainly from the randomness of their input encoding rather than from their spiking computation, and an adversary who repeats each query can largely overcome it. SNNs therefore protect against model stealing without additional defense mechanisms, but this protection does not withstand adaptive adversaries.
\end{abstract}


\section{Introduction}
\label{sec:intro}

Machine learning models increasingly run on edge devices~\cite{zhou2019edge} in sensitive domains such as healthcare~\cite{price2019privacy} and finance~\cite{bussmann2021machine}. Deployment exposes these models to model stealing attacks (MSAs), in which an adversary queries a model, observes its predictions, and trains a surrogate that replicates its functionality using only the top-1 labels that any deployed classifier returns~\cite{tramer2016stealing, papernot2017practical, orekondy2019knockoff, correiasilva2018copycat}. The surrogate transfers the value of the model to the adversary and serves as a white-box proxy for further attacks, such as transferable adversarial examples~\cite{papernot2016transferability, papernot2017practical} and membership inference attacks (MIAs)~\cite{jagielski2020high}.

Existing defenses detect suspicious query patterns~\cite{juuti2019prada} or perturb the returned predictions~\cite{orekondy2020prediction}. Both add a mechanism to the deployed system, which is costly on edge devices with limited energy and memory, and an adversary who owns the device can query it without the monitoring that a cloud service enforces. These constraints motivate architectures that limit extraction inherently.

Spiking neural networks (SNNs) are a promising candidate for such an architecture. SNNs communicate through sparse binary spikes and integrate their inputs over time in leaky integrate-and-fire (LIF) neurons, which enables energy-efficient inference on neuromorphic hardware~\cite{roy2019towards}. Prior work shows that, compared with artificial neural networks (ANNs), SNNs reduce the success of MIAs~\cite{moshruba2025neuromorphic, 11228352, 11345916, 11412839} and model inversion~\cite{poursiami2024brainleaks, poursiami2025spikes}, and that rate-encoded SNNs resist black-box adversarial examples~\cite{sharmin2020inherent}. These studies, however, address data-centric threats or input perturbations, and the resistance of SNNs to the replication of their decision function has not been systematically examined.

The stochasticity of rate encoding complicates this evaluation. Rate encoding samples spikes at random, so a rate-encoded SNN can return different labels for repeated queries of the same input. Lower agreement with an SNN target may therefore reflect noise in its outputs rather than a decision function that is harder to approximate. Standard extraction metrics assume a deterministic target and conflate these two effects, yet separating them determines whether a defender can rely on the resistance of SNNs.

This paper examines whether SNNs limit model extraction in the absence of dedicated defense mechanisms and identifies the mechanism responsible for any such resistance. We compare ANN and rate-encoded SNN targets with a shared convolutional backbone under four label-only attacks, namely Papernot~\cite{papernot2017practical}, N I-FGSM, T-RND I-FGSM, and COLOR~\cite{juuti2019prada}, on MNIST~\cite{lecun1998gradient}, Fashion-MNIST~\cite{xiao2017fashion}, and CIFAR-10~\cite{krizhevsky2009learning}, and we measure fidelity, adversarial transferability, and query efficiency. To separate the fidelity of the surrogate from the stochasticity of the target, we normalize agreement by the output consistency of the SNN target, which bounds the agreement that any surrogate can attain. We also evaluate an adaptive adversary that queries each input five times and keeps the most frequent label, which we refer to as majority voting.

Under single-query attacks, the SNN target resists extraction more strongly than the ANN target. Surrogates match its predictions on natural inputs in most settings, but their agreement with it on random inputs is 38\% lower on average, and targeted adversarial examples crafted on them transfer to it 54\% less often. Most of this resistance, however, stems from stochastic encoding: normalization by output consistency reduces the deficit in random agreement to 8\%, and majority voting restores targeted transfer on MNIST to the level attained against the ANN target. Rate-encoded SNNs therefore resist extraction without additional mechanisms, but an adaptive adversary can remove most of this resistance at a higher query cost.

The main contributions of this paper are as follows:
\begin{itemize}
    \item \textbf{A systematic benchmark of model extraction against SNNs.} We evaluate four label-only attacks against ANN and SNN targets on three datasets in terms of fidelity, adversarial transferability, and query efficiency.
    \item \textbf{A methodology for evaluating extraction against stochastic targets.} We introduce output consistency, which bounds the agreement that any surrogate can attain and separates the fidelity of the surrogate from the noise of the target.
    \item \textbf{An attribution of SNN resistance to stochastic encoding.} We show that stochastic outputs account for most of the reduction in random agreement and adversarial transferability that SNN targets exhibit.
    \item \textbf{An evaluation of adaptive adversaries and query budgets.} We show that majority voting restores targeted transferability to the level of the ANN target, whereas a larger single-query budget does not raise agreement beyond the consistency bound; only a 6\% gap in test agreement on MNIST persists.
\end{itemize}

\section{Background}
\label{sec:background}
 
Privacy attacks on machine learning models span several threat models. Model inversion attacks recover representative inputs or sensitive attributes from model predictions~\cite{fredrikson2015model}, and MIAs determine whether specific samples were used during training~\cite{shokri2017membership, yeom2018privacy}. These attacks succeed against a range of ANN architectures and show that black-box query access alone can expose information about the training data of a model. For SNNs, prior work reports reduced vulnerability under both threat models, with lower MIA success rates~\cite{moshruba2025neuromorphic, 11228352, 11345916, 11412839} and greater resistance to input reconstruction~\cite{poursiami2024brainleaks, poursiami2025spikes}. These studies generally attribute this reduction to the sparse, event-driven nature of spike-based computation and the temporal dynamics of LIF neurons. A related line of work finds that SNNs with Poisson rate encoding resist adversarial examples better than ANNs, particularly in black-box settings, and attributes this robustness to the discretization that the encoder introduces and to the leak of LIF neurons~\cite{sharmin2020inherent}.
 
Existing privacy studies of SNNs, however, address data-centric threats, in which the adversary seeks information about training samples. Whether the same properties limit the ability of an adversary to replicate the decision function of a model through query access has not been systematically examined; work on watermarking SNNs evaluates extraction only as a means of evading ownership verification~\cite{riano2026memark}. We address this question by comparing the extractability of ANN and SNN targets under identical attacks and query budgets.

\subsection{Spiking Neural Networks}
\label{sec:snn_background}
 
SNNs encode information through discrete spike events over time rather than continuous activations~\cite{maass1997networks, gerstner2002spiking}. In the discrete-time LIF model~\cite{gerstner2002spiking, burkitt2006review}, the membrane potential of a neuron evolves as
\begin{align}
V_t &= \beta V_{t-1} + I_t - V_{th} S_{t-1}, \label{eq:lif} \\
S_t &= \Theta(V_t - V_{th}), \label{eq:spike}
\end{align}
where $I_t$ denotes the input current, $\beta \in (0, 1)$ the leakage factor, $V_{th}$ the threshold, $S_t \in \{0, 1\}$ the spike output, and $\Theta$ the Heaviside step function. The last term of~\eqref{eq:lif} implements reset by subtraction. Because $\Theta$ has zero gradient almost everywhere, SNNs are trained with backpropagation through time and surrogate gradients~\cite{neftci2019surrogate}, and the prediction is read out from the spike counts or the membrane potential of the output layer.
 
\paragraph{Input Encoding}
Static inputs must be converted into spike trains over $T$ time steps. Rate encoding samples a spike for each input element at every step from a Bernoulli distribution whose probability equals the normalized intensity of that element~\cite{eshraghian2023training}. Because the encoder draws a new spike train in every forward pass, repeated queries of the same input can return different labels. Direct encoding instead supplies the analog input at every step and yields deterministic outputs~\cite{rathi2023diet}.
 
\paragraph{Implications for Model Extraction}
Two properties of SNNs could therefore limit model extraction. Temporal dynamics may yield a decision function that an ANN surrogate approximates less accurately, and stochastic encoding returns noisy labels that corrupt the training data of the surrogate and limit the agreement that any surrogate can attain (Section~\ref{sec:metrics}). Our evaluation distinguishes between these two explanations (Section~\ref{sec:results}).

\subsection{Model Stealing Attacks}
\label{sec:msa_background}

MSAs replicate the functionality of a target model through black-box query access~\cite{tramer2016stealing}. Jagielski et al.~\cite{jagielski2020high} distinguish accuracy, which measures how well the surrogate performs the original task, from fidelity, which measures how closely it reproduces the predictions of the target; we evaluate fidelity. We consider four attacks: Papernot~\cite{papernot2017practical} and three attacks from the taxonomy of Juuti et al.~\cite{juuti2019prada}, namely N I-FGSM, T-RND I-FGSM, and COLOR.

\begin{figure*}[t]
    \centering
    \includegraphics[width=\linewidth]{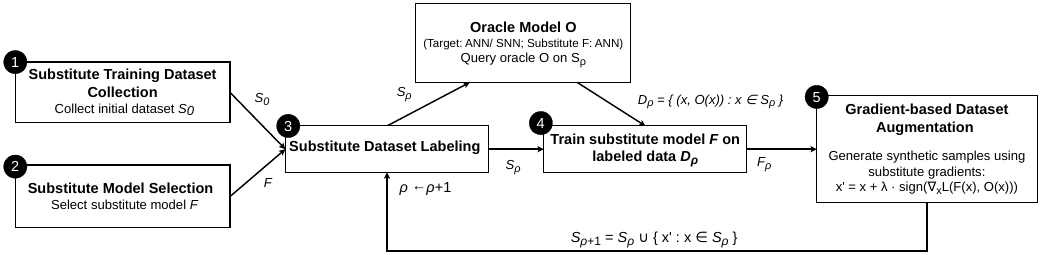}
    \caption{Iterative extraction procedure, illustrated for the Papernot attack. In each round $\rho$, the adversary queries the target $O$ for the labels of the input set $S_\rho$, trains the surrogate $F$ on the labeled set $D_\rho$, synthesizes a new input $x'$ from each $x \in S_\rho$ with~\eqref{eq:papernot}, and updates the input set as $S_{\rho+1} = S_\rho \cup \{x' \mid x \in S_\rho\}$ until the query budget is exhausted.}
    \label{fig:papernot}
\end{figure*}

All four attacks share an iterative procedure (Fig.~\ref{fig:papernot}). We denote the target, also called the oracle, by $O$ and the surrogate, also called the substitute, by $F$; $O(x)$ denotes the label that the target returns for input $x$. Starting from a small seed set $S_0$ of natural inputs, the adversary in each round $\rho$ trains $F$ on the labeled set $D_\rho = \{(x, O(x)) \mid x \in S_\rho\}$, synthesizes new inputs $S'_\rho$ from $S_\rho$, queries the target once for their labels, and sets $S_{\rho+1} = S_\rho \cup S'_\rho$ until the query budget is exhausted. Because the adversary cannot compute gradients of $O$, the gradient-based attacks compute them on $F$. The attacks differ only in how they synthesize $S'_\rho$.

\subsubsection{Papernot}
Papernot et al.~\cite{papernot2017practical} perturb each sample by a single step along the sign of the input gradient of the surrogate,
\begin{equation}
x' = x + \lambda \cdot \mathrm{sign}\big(\nabla_x \mathcal{L}(F(x), O(x))\big),
\label{eq:papernot}
\end{equation}
where $\mathcal{L}$ denotes the cross-entropy loss and $\lambda$ the step size, so that new queries probe the target where the predictions of the surrogate change. Each round doubles the labeled set.

\subsubsection{N I-FGSM}
Tram\`er et al.~\cite{tramer2016stealing} retrained the surrogate on queries near the decision boundary of the target, which they located through line search. N I-FGSM~\cite{juuti2019prada} pursues the same goal with the non-targeted iterative fast gradient sign method (I-FGSM). Starting from $x^{(0)} = x$, each of $n$ iterations increases the loss of the surrogate for the label of the target,
\begin{equation}
x^{(j+1)} = \Pi_{\epsilon}\Big(x^{(j)} + \alpha \cdot \mathrm{sign}\big(\nabla_x \mathcal{L}(F(x^{(j)}), O(x))\big)\Big),
\label{eq:nifgsm}
\end{equation}
where $\alpha$ denotes the step size and $\Pi_{\epsilon}$ projects onto the $L_\infty$ ball of radius $\epsilon$ around $x$. The synthesized input is $x' = x^{(n)}$. Like Papernot, N I-FGSM synthesizes one input per sample in each round.

\subsubsection{T-RND I-FGSM}
T-RND I-FGSM~\cite{juuti2019prada} draws a target class $y^{\text{tgt}}$ uniformly from all classes except $O(x)$ and replaces the ascent step in~\eqref{eq:nifgsm} with a descent step on $\mathcal{L}(F(x^{(j)}), y^{\text{tgt}})$. It thereby directs queries toward the decision regions of all classes rather than toward the boundary nearest to each sample.

\subsubsection{COLOR}
\begin{figure}[t]
    \centering
    \includegraphics[width=\linewidth]{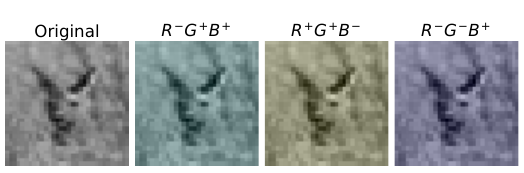}
    \caption{COLOR perturbations of a CIFAR-10 test image. Each variant adds an offset of $+\lambda$ or $-\lambda$ ($\lambda = 25.5/255$ in the normalized input space) to every pixel of a color channel, with the sign drawn independently for each channel; panel titles give the signs. Images are shown at the true perturbation magnitude.}
    \label{fig:color_examples}
\end{figure}

COLOR~\cite{juuti2019prada} synthesizes queries without gradients by adding an offset $\lambda s_c$ to every pixel of each channel $c$, where each sign $s_c \in \{-1, +1\}$ is drawn uniformly at random. For grayscale inputs, the offset changes only the brightness of the image; for color inputs, independent signs also shift its color balance (Fig.~\ref{fig:color_examples}). COLOR therefore serves as a gradient-free baseline.

\section{Experimental Setup}
\label{sec:setup}

We compare the extractability of ANN and SNN targets that share the same datasets, convolutional backbone, training schedule, and attack configurations.

\subsection{Datasets and Models}
\label{sec:models}
 
\paragraph{Datasets}
We evaluate on MNIST~\cite{lecun1998gradient} and Fashion-MNIST~\cite{xiao2017fashion}, each with 60{,}000 training and 10{,}000 test grayscale images of $28 \times 28$ pixels, and on CIFAR-10~\cite{krizhevsky2009learning}, with 50{,}000 training and 10{,}000 test color images of $32 \times 32$ pixels. MNIST and Fashion-MNIST pixel values lie in $[0, 1]$, and we normalize CIFAR-10 images to $[-1, 1]$. CIFAR-10 training uses random $32 \times 32$ crops with a padding of 4 pixels and random horizontal flips.
 
\paragraph{Architectures}
Both targets share the convolutional backbone in Table~\ref{tab:architectures}, with two convolutional layers followed by two fully connected (FC) layers. The ANN target predicts the argmax of its output logits. The SNN target processes each input over $T = 10$ time steps. At every step, a rate encoder samples a Bernoulli spike for each pixel with probability $p = \mathrm{clip}(x, 0, 1)$ for MNIST and Fashion-MNIST and $p = \mathrm{clip}((x + 1)/2, 0, 1)$ for CIFAR-10. The convolutional layers process these spikes with ReLU activations, and LIF neurons (Section~\ref{sec:snn_background}) with $\beta = 0.95$, $V_{th} = 1.0$, and reset by subtraction follow both FC layers. The SNN target predicts the argmax of the membrane potential of the output layer at the final time step. The surrogate uses the ANN architecture and therefore matches the ANN target exactly, whereas it differs from the SNN target in input encoding and neuron model.
 
\begin{table}[t]
\caption{Architectures of the targets. Each convolutional layer uses a padding of 1 and $2 \times 2$ max pooling. The first FC layer receives $64 \times 7 \times 7$ features for MNIST and Fashion-MNIST and $64 \times 8 \times 8$ for CIFAR-10. The surrogate uses the ANN architecture.}
\label{tab:architectures}
\centering
\small
\begin{tabular}{lll}
\toprule
Layer & ANN & SNN \\
\midrule
Input encoding & None & Rate, $T = 10$ \\
Conv, 32 filters, $3 \times 3$ & ReLU & ReLU \\
Conv, 64 filters, $3 \times 3$ & ReLU & ReLU \\
FC, 1{,}000 units & ReLU & LIF \\
FC, 10 units & None & LIF \\
Prediction & Logits & Membrane potential \\
\bottomrule
\end{tabular}
\end{table}
 
\paragraph{Target Training}
We train each target from scratch with cross-entropy loss and Adam (learning rate $10^{-3}$, batch size 128, no weight decay) for 40 epochs on MNIST and Fashion-MNIST and 100 epochs on CIFAR-10, and retain the checkpoint with the highest test accuracy. The SNN target computes the loss on the final membrane potential of the output layer and uses the default arctangent surrogate gradient of snnTorch~\cite{eshraghian2023training}. Table~\ref{tab:target_properties} reports the test accuracy of each target.

\subsection{Attack Setup}
\label{sec:attack_setup}
 
\paragraph{Threat Model}
The adversary observes only the top-1 label of the target, without access to confidence scores, parameters, or architecture, and holds a budget of 102{,}400 extraction queries. Every extraction query counts toward this budget, including repeated queries of the same input; the queries that we issue to evaluate a finished surrogate do not.
 
\paragraph{Extraction Procedure}
Each attack starts from a balanced seed set of 10 samples per class drawn from the training split, which can overlap with the training data of both targets. We initialize the surrogate once and fine-tune it in every round for 10 epochs with Adam (learning rate $10^{-3}$, batch size 128), reinitializing the optimizer state each round. The attack stops after 10 rounds or when it exhausts the budget.
 
\paragraph{Query Synthesis}
Papernot and COLOR use $\lambda = 25.5/255$. N I-FGSM uses $\epsilon = 8/255$ with $n = 5$ iterations, and T-RND I-FGSM uses $\epsilon = 64/255$ with $n = 11$; both use the step size $\alpha = \epsilon / n$. All attacks clip queries to $[-1, 1]$. For MNIST and Fashion-MNIST, whose inputs lie in $[0, 1]$, this range admits negative values, which the SNN encoder clamps to 0 but the ANN surrogate processes unchanged. For CIFAR-10, the encoder maps $[-1, 1]$ to $[0, 1]$ and preserves the full range. Papernot and N I-FGSM synthesize one query per labeled sample and reach the budget after 10 rounds ($100 \times 2^{10}$). T-RND I-FGSM synthesizes nine variants and COLOR four variants of each sample added in the previous round, and both fill the remaining budget in their final round.
 
\paragraph{Adaptive Attacker}
We also evaluate a majority-vote variant of Papernot on MNIST that queries each input five times and keeps the most frequent label. Because all queries count toward the budget, it labels at most 20{,}480 distinct inputs.
 
\paragraph{Transferability and Repetitions}
We craft adversarial examples on the finished surrogate with targeted and non-targeted I-FGSM ($\epsilon = 64/255$, $n = 11$), following Juuti et al.~\cite{juuti2019prada}. We repeat each configuration with three random seeds and report the mean. The random seed determines the initial seed set, the initialization of the surrogate, the order of its training data, and the spike sampling of the SNN target, whereas the target checkpoints remain fixed.

\subsection{Evaluation Metrics}
\label{sec:metrics}
 
We evaluate extraction by fidelity, adversarial transferability, and query efficiency. As in Section~\ref{sec:msa_background}, $O$ denotes the target and $F$ the surrogate; here, $F(x)$ denotes the class that the surrogate predicts. The SNN target predicts the argmax of its output membrane potential at the final time step, and $O(x)$ denotes the label that a single query returns. We evaluate every metric with one query per input, as an attacker or a user of the deployed model would observe, and $\mathds{1}(\cdot)$ denotes the indicator function.
 
\paragraph{Fidelity}
Agreement on an evaluation set $\mathcal{D} = \{x_i\}_{i=1}^{N}$ is
\begin{equation}
\text{Agr}(\mathcal{D}) = \frac{1}{N} \sum_{i=1}^{N} \mathds{1}\big(O(x_i) = F(x_i)\big).
\end{equation}
Test agreement uses the full test split (10{,}000 inputs) and measures in-distribution fidelity. Random agreement uses 4{,}000 inputs sampled uniformly from $[-1, 1]$ with a fixed seed, which lie outside the data distribution.
 
\paragraph{Adversarial Transferability}
We craft adversarial examples $x_i'$ on the surrogate for the first 1{,}000 test inputs. Non-targeted transferability (NT) is the fraction of inputs for which the perturbation changes the prediction of the target. Targeted transferability (TT) is the fraction that the target assigns to a class $y_i^{\text{tgt}}$ drawn uniformly from all classes except $O(x_i)$:
\begin{align}
\text{NT} &= \frac{1}{N} \sum_{i=1}^{N} \mathds{1}\big(O(x_i') \neq O(x_i)\big), \\
\text{TT} &= \frac{1}{N} \sum_{i=1}^{N} \mathds{1}\big(O(x_i') = y_i^{\text{tgt}}\big).
\end{align}
TT additionally requires the surrogate to capture which class lies beyond each decision boundary. Because the SNN target returns stochastic labels, single-query evaluation may overstate its NT.
 
\paragraph{Query Efficiency}
After the seed stage and each round, we record the cumulative number of queries and the test and random agreement of the current surrogate. We do this for Papernot and T-RND I-FGSM in three runs on MNIST and in one run on Fashion-MNIST and CIFAR-10, separately from the main runs. Each run labels the evaluation sets once, so that changes across checkpoints reflect only the surrogate.
 
\paragraph{Output Consistency}
Let $p_c(x)$ denote the probability that a single query of the SNN target returns class $c$. Any surrogate agrees with a single query with probability at most $\max_c p_c(x)$, which it attains by predicting the most probable class. We define the output consistency of the target as the average of this maximum over an evaluation set and estimate it from $K = 25$ queries per input. With $O^{(k)}(x_i)$ the label of the $k$-th query and $\hat{y}(x_i)$ the most frequent label among the $K$ queries,
\begin{equation}
C(\mathcal{D}) = \frac{1}{NK} \sum_{i=1}^{N} \sum_{k=1}^{K} \mathds{1}\big(O^{(k)}(x_i) = \hat{y}(x_i)\big).
\end{equation}
Output consistency thus estimates the highest expected agreement that any surrogate can attain under single-query evaluation. Because $\hat{y}$ and the agreement derive from the same queries, the estimate slightly overstates this bound when several classes are nearly equally probable. The deterministic ANN target has an output consistency of 1.0. We use output consistency to interpret agreement (Section~\ref{sec:fidelity}), not as a measure of attack success.

\section{Evaluation}
\label{sec:results}

This section compares the extractability of the ANN and SNN targets. Section~\ref{sec:fidelity} reports fidelity, Section~\ref{sec:transferability} adversarial transferability, and Section~\ref{sec:query_budget} the effect of the query budget and of an adaptive adversary. All results use the ANN surrogate described in Section~\ref{sec:attack_setup}.

\subsection{Fidelity of Model Extraction}
\label{sec:fidelity}

Table~\ref{tab:target_properties} summarizes the two targets. Their accuracy is comparable on MNIST and Fashion-MNIST, whereas the SNN target is 7.7 percentage points less accurate on CIFAR-10. The ANN target is deterministic. The SNN target is highly consistent on the test inputs of MNIST (0.997) and Fashion-MNIST (0.959), less consistent on those of CIFAR-10 (0.809), and substantially less consistent on uniform random inputs (0.636, 0.839, and 0.547, respectively).

\begin{table}[t]
\caption{Test accuracy of the targets and output consistency of the SNN target on test and uniform random inputs. The ANN target has an output consistency of 1.0.}
\label{tab:target_properties}
\centering
\small
\begin{tabular}{lcccc}
\toprule
 & \multicolumn{2}{c}{Accuracy (\%)} & \multicolumn{2}{c}{SNN consistency} \\
\cmidrule(lr){2-3} \cmidrule(lr){4-5}
Dataset & ANN & SNN & Test & Uniform \\
\midrule
MNIST & 98.49 & 99.22 & 0.997 & 0.636 \\
Fashion-MNIST & 92.51 & 90.25 & 0.959 & 0.839 \\
CIFAR-10 & 80.21 & 72.56 & 0.809 & 0.547 \\
\bottomrule
\end{tabular}
\end{table}

\begin{figure*}[t]
    \centering
    \includegraphics[width=\linewidth]{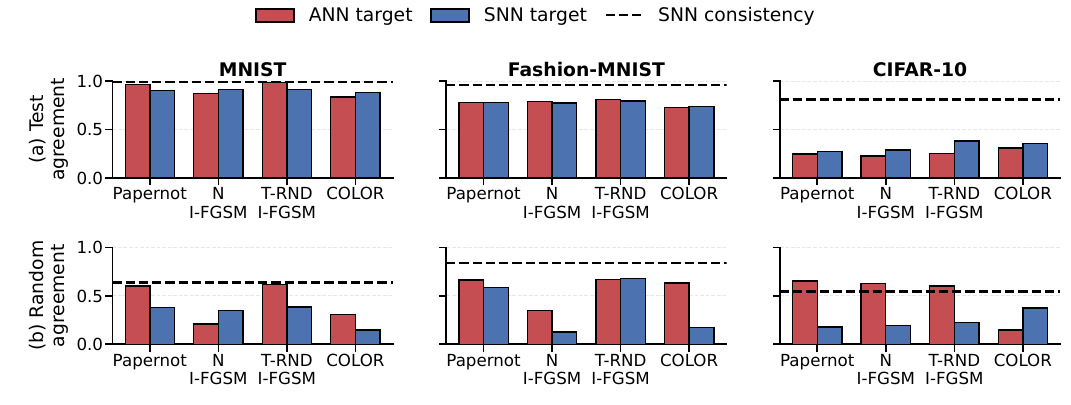}
    \caption{Fidelity of model extraction with an ANN surrogate: (a)~test agreement and (b)~random agreement, as means over three runs. The dashed line denotes the output consistency of the SNN target, the highest agreement that any surrogate can attain against it; it applies only to the SNN bars, and the bound for the ANN target is 1.0.}
    \label{fig:fidelity}
\end{figure*}

Because output consistency bounds the agreement that any surrogate can attain against the SNN target (Section~\ref{sec:metrics}), Fig.~\ref{fig:fidelity} marks it with a dashed line. The line applies only to the SNN bars, and the gap between an SNN bar and the line measures the behavior of the SNN target that the surrogate fails to reproduce. The corresponding bound for the deterministic ANN target is 1.0, so ANN bars above the line do not violate any bound.

\subsubsection{Test Agreement}
On MNIST and Fashion-MNIST, surrogates attain comparable test agreement against both targets (Fig.~\ref{fig:fidelity}(a)). SNN test agreement falls more than 0.02 below ANN test agreement in only 2 of the 12 combinations, MNIST under Papernot (0.903 versus 0.961) and under T-RND I-FGSM (0.913 versus 0.984), and exceeds it in six (Table~\ref{tab:summary}). On CIFAR-10, extraction is ineffective against both targets, with test agreement between 0.23 and 0.38, well below the consistency bound of 0.809. The SNN target yields higher agreement there, which we do not interpret as greater vulnerability because its lower accuracy may make its decision function easier to approximate. Except on MNIST, the SNN target therefore confers no protection on in-distribution data.

\subsubsection{Random Agreement}
On random inputs, the SNN target yields lower agreement in 9 of the 12 combinations, by 0.191 on average (Fig.~\ref{fig:fidelity}(b)). This comparison conflates the fidelity of the surrogate with the stochasticity of the target, whose consistency bound on these inputs is as low as 0.547. We therefore normalize each SNN value by the bound of its dataset, which expresses it as a fraction of the maximum attainable agreement and makes it directly comparable to ANN agreement. After normalization, the SNN target yields lower agreement in six combinations, comparable agreement in two, and higher agreement in four, and the average deficit shrinks to 0.039 (Table~\ref{tab:summary}; Appendix Table~\ref{tab:full_results}). Stochastic outputs thus account for most of the raw difference.

The residual difference concentrates on CIFAR-10, where normalized SNN agreement remains well below ANN agreement under Papernot (0.328 versus 0.654), N I-FGSM (0.354 versus 0.629), and T-RND I-FGSM (0.412 versus 0.603). On MNIST and Fashion-MNIST, the gradient-based attacks show no consistent trend. Random agreement also varies by up to 0.18 between individual runs (Section~\ref{sec:budget_curves}). This variation exceeds the smaller differences on MNIST and Fashion-MNIST but not the CIFAR-10 residuals of 0.19 to 0.33.

The COLOR results on MNIST and Fashion-MNIST require qualification. COLOR queries and uniform inputs contain negative values that the SNN encoder clamps to 0 but the ANN surrogate processes (Section~\ref{sec:attack_setup}), so a negative COLOR offset leaves the image background unchanged for the SNN target. We attribute the large COLOR differences on these datasets (normalized agreement of 0.231 versus 0.309 on MNIST and 0.204 versus 0.632 on Fashion-MNIST) at least in part to this mismatch. The mismatch also affects random agreement under every attack on these two datasets but does not arise on CIFAR-10.

In summary, the resistance of the SNN target outside the data distribution derives primarily from its stochastic outputs. A residual advantage remains only on CIFAR-10, where extraction is ineffective against both targets. Section~\ref{sec:majority} evaluates an adversary that mitigates this stochasticity through repeated queries.

\subsection{Adversarial Transferability}
\label{sec:transferability}

\begin{figure*}[t]
    \centering
    \includegraphics[width=\linewidth]{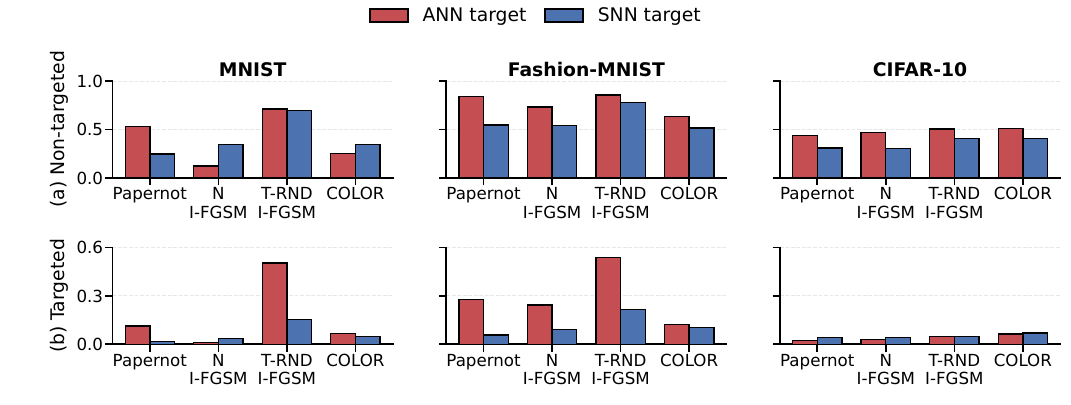}
    \caption{Adversarial transferability from the ANN surrogate to each target: (a)~NT and (b)~TT for I-FGSM adversarial examples ($\epsilon = 64/255$, 11 iterations) on the first 1{,}000 test examples, as means over three runs.}
    \label{fig:transferability}
\end{figure*}

Transferability measures whether the surrogate inherits the vulnerabilities of the target (Fig.~\ref{fig:transferability}). NT is lower for the SNN target in 9 of the 12 combinations, by 0.097 on average. The reduction is consistent on Fashion-MNIST (0.076 to 0.296) and CIFAR-10 (0.100 to 0.166). MNIST is mixed: NT decreases under Papernot (from 0.527 to 0.245), remains comparable under T-RND I-FGSM, and increases under N I-FGSM and COLOR, possibly because single-query evaluation counts stochastic label changes as transfer.

\begin{figure*}[t]
    \centering
    \includegraphics[width=\linewidth]{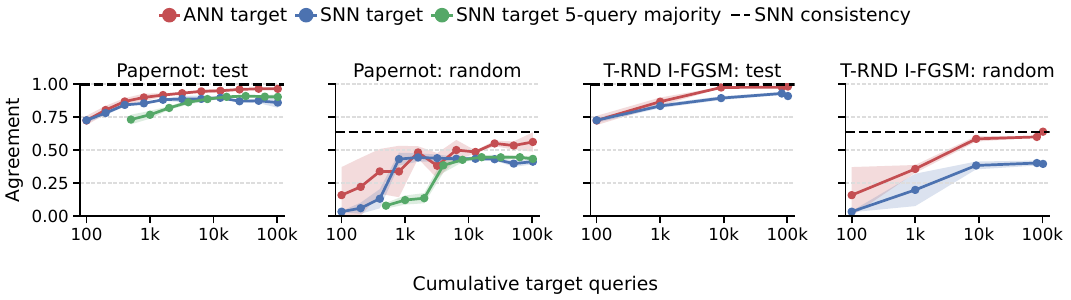}
    \caption{Test and random agreement on MNIST as a function of the cumulative number of target queries. Lines denote means and bands one standard deviation over three runs; green curves denote the majority-vote attacker, whose first checkpoint lies at 500 queries. The dashed line denotes the output consistency of the SNN target. Each run labels the evaluation sets once, so final values differ slightly from those in Tables~\ref{tab:majority} and~\ref{tab:full_results}.}
    \label{fig:query_budget}
\end{figure*}

TT reveals the largest difference between the targets, because success requires the surrogate to capture the class beyond each boundary. Under T-RND I-FGSM, the most effective targeted attack, TT reaches 0.503 against the ANN target on MNIST but only 0.154 against the SNN target; on Fashion-MNIST, the corresponding values are 0.537 and 0.215. Papernot shows the same trend on both datasets. On CIFAR-10, TT does not exceed 0.07 for either target. Overall, the SNN target yields lower TT in five combinations, comparable TT in six, and higher TT in one.

\begin{table}[t]
\caption{Comparison of the SNN and ANN targets across the 12 combinations of attack and dataset. Lower, Comp., and Higher count the combinations in which the SNN value is lower than, comparable to (difference of at most 0.02), or higher than the ANN value; Mean diff.\ is the mean difference (SNN minus ANN). Random agreement varies by up to 0.18 between runs, so its counts are indicative only.}
\label{tab:summary}
\centering
\small
\setlength{\tabcolsep}{3pt}
\begin{tabular}{lcccc}
\toprule
Metric & Lower & Comp. & Higher & Mean diff. \\
\midrule
Test agreement & 2 & 4 & 6 & $+0.017$ \\
Random agreement & 9 & 1 & 2 & $-0.191$ \\
Random (normalized) & 6 & 2 & 4 & $-0.039$ \\
NT & 9 & 1 & 2 & $-0.097$ \\
TT & 5 & 6 & 1 & $-0.092$ \\
\bottomrule
\end{tabular}
\end{table}

Under single-query attacks, reduced transferability is thus the most consistent distinction between the targets: surrogates reproduce the labels of the SNN target on natural inputs as accurately as those of the ANN target, yet their adversarial examples transfer to it less often. Section~\ref{sec:majority} examines whether this distinction persists when the adversary reduces label noise.

\subsection{Query Budget and Adaptive Attackers}
\label{sec:query_budget}

We now examine how fidelity develops as the adversary spends the budget and whether an adversary that repeats queries recovers what single-query attacks fail to extract. We focus on MNIST, where the SNN target yields the largest reduction in test agreement.

\subsubsection{Fidelity across the Query Budget}
\label{sec:budget_curves}
Fig.~\ref{fig:query_budget} reports test and random agreement on MNIST after each round of Papernot and T-RND I-FGSM over three runs. These runs are separate from those of Sections~\ref{sec:fidelity} and~\ref{sec:transferability}, and their final values differ from them by at most 0.044. Appendix Fig.~\ref{fig:query_budget_appendix} reports single-run curves for Fashion-MNIST and CIFAR-10, in which the two targets yield similar test agreement on Fashion-MNIST and the SNN target yields slightly higher test agreement on CIFAR-10.

Both surrogates start from the same test agreement (0.724 and 0.725) but diverge from the first round. Against the ANN target, test agreement rises steadily to 0.963 under Papernot and 0.981 under T-RND I-FGSM. Against the SNN target, it saturates early. Under Papernot, the mean peaks at 0.896 after 12{,}800 queries, below the 0.898 that the ANN surrogate attains after 800 queries; under T-RND I-FGSM, it peaks at 0.928. Because output consistency on test inputs is 0.997, label noise on natural inputs cannot explain this ceiling. Under Papernot, the mean also declines after 12{,}800 queries, to 0.859, in two of the three runs. This decline is consistent with noisy labels accumulating as synthesized queries move away from the data distribution.

Random agreement against the SNN target rises quickly and then plateaus between 0.40 and 0.44 under Papernot and between 0.38 and 0.40 under T-RND I-FGSM. These plateaus correspond to 0.60 to 0.70 of the consistency bound, comparable to or above the 0.56 to 0.64 that the ANN surrogate attains of its bound. A larger budget therefore does not close the gap on MNIST. Off the data distribution, the surrogate reaches the limit imposed by output consistency within a few thousand queries; on the data distribution, test agreement saturates below the level that the ANN surrogate attains with 800 queries.

\subsubsection{Majority-Vote Attacker}
\label{sec:majority}
Section~\ref{sec:fidelity} attributes most of the lower random agreement to stochastic outputs, which an adaptive adversary can suppress through majority voting (Section~\ref{sec:attack_setup}). Table~\ref{tab:majority} compares this attacker with the single-query attacks. Majority voting leaves test agreement unchanged (0.902) and raises normalized random agreement only from 0.595 to 0.658. In Fig.~\ref{fig:query_budget}, it reaches a random agreement plateau similar to that of the single-query attacker but requires ten times as many queries. Its test agreement does not decline in late rounds, which supports the attribution of the single-query decline to noisy labels.

\begin{table}[t]
\caption{Single-query and majority-vote extraction of the MNIST targets under Papernot, as means over three runs. Rand.\ denotes random agreement, and Norm.\ divides it by the output consistency of the target (0.636 for the SNN target and 1.0 for the ANN target). Standard deviations of the majority-vote attacker are 0.018, 0.015, 0.017, and 0.025 for test agreement, random agreement, NT, and TT.}
\label{tab:majority}
\centering
\small
\setlength{\tabcolsep}{3pt}
\begin{tabular}{lccccc}
\toprule
Target, attacker & Test & Rand. & Norm. & NT & TT \\
\midrule
ANN, single query & 0.961 & 0.600 & 0.600 & 0.527 & 0.114 \\
SNN, single query & 0.903 & 0.378 & 0.595 & 0.245 & 0.019 \\
SNN, 5-query majority & 0.902 & 0.418 & 0.658 & 0.636 & 0.120 \\
\bottomrule
\end{tabular}
\end{table}

The effect of majority voting on transferability, in contrast, is substantial. NT rises from 0.245 to 0.636, above the ANN level of 0.527, partly because single-query evaluation counts stochastic label changes as transfer. TT rises from 0.019 to 0.120, matching the ANN level of 0.114. On MNIST under Papernot, the reduced transferability of the SNN target therefore derives primarily from label noise in the extraction data, which an adversary removes at five times the query cost per label. The lower test agreement, in contrast, persists (0.902 versus 0.961). This gap is confined to MNIST, as Fashion-MNIST and CIFAR-10 exhibit no corresponding reduction in test agreement. It therefore does not constitute a general property of SNN extraction, and its source remains a subject for future investigation.

\section{Conclusion}
\label{sec:conclusion}

The value of an architecture as a defense against MSAs depends not only on whether it resists extraction but also on why it does. In this work, we addressed both questions for SNNs under four label-only attacks on three datasets. Rate-encoded SNN targets reduced agreement on random inputs and the transferability of adversarial examples, particularly targeted transfer, while surrogates reproduced their predictions on natural inputs as faithfully as those of ANN targets in most settings. By bounding the attainable agreement with output consistency and evaluating an adversary that applies majority voting over repeated queries, we traced most of this resistance to the stochasticity of rate encoding rather than to a decision function that is harder to approximate. Rate encoding therefore acts as an implicit form of output randomization: it raises the cost of extraction without an additional mechanism, yet an adaptive adversary can remove most of its effect by repeating queries. These findings indicate that the resistance of SNNs to extraction depends substantially on how inputs are encoded, not only on spiking computation, and that evaluations of stochastic models should report output consistency alongside agreement to avoid attributing output noise to the robustness of the model.

\bibliographystyle{IEEEtran}
\bibliography{sample-base}

\onecolumn
\appendix
\section{Additional Results}
\label{sec:appendix}

\begin{figure}[h!]
    \centering
    \includegraphics[width=0.9\linewidth]{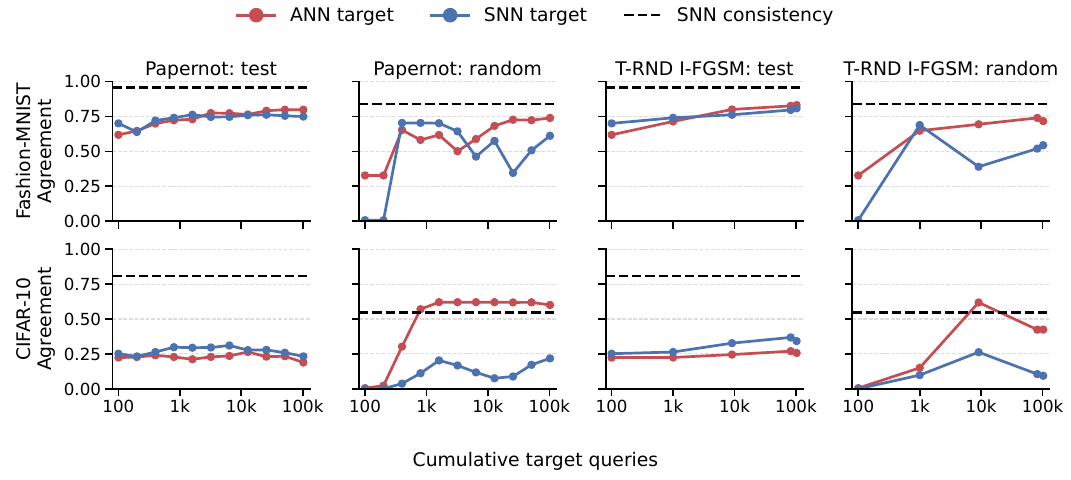}
    \caption{Test and random agreement as a function of the cumulative number of target queries on Fashion-MNIST (top) and CIFAR-10 (bottom) under the Papernot and T-RND I-FGSM attacks, from a single run. The dashed line denotes the output consistency of the SNN target and applies only to the SNN curves.}
    \label{fig:query_budget_appendix}
\end{figure}

Table~\ref{tab:full_results} reports the extraction results for every combination of attack and dataset that Sections~\ref{sec:fidelity} and~\ref{sec:transferability} summarize. Figure~\ref{fig:query_budget_appendix} reports fidelity across the query budget on Fashion-MNIST and CIFAR-10 from a single run and complements the MNIST results in Section~\ref{sec:budget_curves}.

\begin{table}[h!]
\caption{Extraction results for the ANN and SNN targets with an ANN surrogate. Values are means over three runs. Normalized random agreement divides SNN random agreement by the output consistency of the SNN target on uniform inputs (Table~\ref{tab:target_properties}); the corresponding bound for the ANN target is 1.0.}
\label{tab:full_results}
\centering
\small
\setlength{\tabcolsep}{4pt}
\begin{tabular}{lccccccccc}
\toprule
 & \multicolumn{2}{c}{Test agreement} & \multicolumn{3}{c}{Random agreement} & \multicolumn{2}{c}{NT} & \multicolumn{2}{c}{TT} \\
\cmidrule(lr){2-3} \cmidrule(lr){4-6} \cmidrule(lr){7-8} \cmidrule(lr){9-10}
Attack & ANN & SNN & ANN & SNN & SNN (norm.) & ANN & SNN & ANN & SNN \\
\midrule
\multicolumn{10}{l}{\textit{MNIST}} \\
\quad Papernot & 0.961 & 0.903 & 0.600 & 0.378 & 0.595 & 0.527 & 0.245 & 0.114 & 0.019 \\
\quad N I-FGSM & 0.870 & 0.909 & 0.211 & 0.352 & 0.554 & 0.120 & 0.342 & 0.010 & 0.037 \\
\quad T-RND I-FGSM & 0.984 & 0.913 & 0.621 & 0.383 & 0.603 & 0.712 & 0.694 & 0.503 & 0.154 \\
\quad COLOR & 0.834 & 0.881 & 0.309 & 0.147 & 0.231 & 0.252 & 0.346 & 0.066 & 0.049 \\
\midrule
\multicolumn{10}{l}{\textit{Fashion-MNIST}} \\
\quad Papernot & 0.775 & 0.778 & 0.663 & 0.586 & 0.698 & 0.840 & 0.544 & 0.279 & 0.057 \\
\quad N I-FGSM & 0.787 & 0.771 & 0.351 & 0.125 & 0.150 & 0.733 & 0.540 & 0.243 & 0.091 \\
\quad T-RND I-FGSM & 0.807 & 0.792 & 0.666 & 0.677 & 0.807 & 0.855 & 0.779 & 0.537 & 0.215 \\
\quad COLOR & 0.728 & 0.735 & 0.632 & 0.171 & 0.204 & 0.635 & 0.514 & 0.121 & 0.105 \\
\midrule
\multicolumn{10}{l}{\textit{CIFAR-10}} \\
\quad Papernot & 0.244 & 0.269 & 0.654 & 0.180 & 0.328 & 0.437 & 0.306 & 0.023 & 0.042 \\
\quad N I-FGSM & 0.226 & 0.289 & 0.629 & 0.194 & 0.354 & 0.467 & 0.301 & 0.028 & 0.042 \\
\quad T-RND I-FGSM & 0.249 & 0.382 & 0.603 & 0.225 & 0.412 & 0.505 & 0.405 & 0.047 & 0.047 \\
\quad COLOR & 0.307 & 0.354 & 0.146 & 0.373 & 0.682 & 0.510 & 0.408 & 0.064 & 0.070 \\
\bottomrule
\end{tabular}
\end{table}

\end{document}